%% file: main.tex
\documentclass[runningheads]{llncs}

\usepackage{eccv}

\usepackage{eccvabbrv}

\usepackage{graphicx}
\usepackage{booktabs}

\usepackage[accsupp]{axessibility}  

\usepackage[pagebackref,breaklinks,colorlinks,citecolor=eccvblue]{hyperref}

\usepackage{orcidlink}

\usepackage{color, colortbl}
\usepackage{graphicx}
\usepackage{tabularx}
\usepackage{multirow}
\usepackage{adjustbox}
\usepackage{bbding}
\usepackage{wrapfig}
\def \OURMETHOD {TGM\xspace}

\usepackage[capitalize]{cleveref}
\crefname{section}{Sec.}{Secs.}
\Crefname{section}{Section}{Sections}
\Crefname{table}{Table}{Tables}
\crefname{table}{Tab.}{Tabs.}
\Crefname{tab}{Table}{Tables}
\crefname{tab}{Tab.}{Tabs.}
\Crefname{figure}{Figure}{Figures}
\crefname{figure}{Fig.}{Figs.}

\begin{document}

\title{Text-Guided Video Masked Autoencoder} 

\titlerunning{Text-Guided Video Masked Autoencoder}

\author{David Fan\inst{1} \and
Jue Wang\inst{1} \and
Shuai Liao\inst{2} \and
Zhikang Zhang\inst{1} \and
Vimal Bhat\inst{1} \and
Xinyu Li\inst{1}
}

\authorrunning{Fan et al.}

\institute{Amazon Prime Video \and Amazon Fulfillment Technology}

\maketitle

\input{0_abstract_cameraready}    
\input{1_intro_cameraready}
\input{2_relatedwork_cameraready}
\input{3_method_cameraready}
\input{4_results_cameraready}
\input{5_conclusion_cameraready}


%
%

\bibliographystyle{splncs04}
\bibliography{main}

\newpage

\appendix
\include{supplementary}

\end{document}

%% file: 0_abstract_cameraready.tex
\begin{abstract}
Recent video masked autoencoder (MAE) works have designed improved masking algorithms focused on saliency. These works leverage visual cues such as motion to mask the most salient regions. However, the robustness of such visual cues depends on how often input videos match underlying assumptions. On the other hand, natural language description is an information dense representation of video that implicitly captures saliency without requiring modality-specific assumptions, and has not been explored yet for video MAE. To this end, we introduce a novel text-guided masking algorithm (TGM) that masks the video regions with highest correspondence to paired captions. Without leveraging any explicit visual cues for saliency, our TGM is competitive with state-of-the-art masking algorithms such as motion-guided masking. To further benefit from the semantics of natural language for masked reconstruction, we next introduce a unified framework for joint MAE and masked video-text contrastive learning. We show that across existing masking algorithms, unifying MAE and masked video-text contrastive learning improves downstream performance compared to pure MAE on a variety of video recognition tasks, especially for linear probe. Within this unified framework, our TGM achieves the best relative performance on five action recognition and one egocentric datasets, highlighting the complementary nature of natural language for masked video modeling.



%

\end{abstract}

%% file: 1_intro_cameraready.tex
\section{Introduction}
\begin{figure}[t]
\begin{subfigure}[t]{.49\linewidth}
    \centering
    \includegraphics[width=0.99\textwidth]{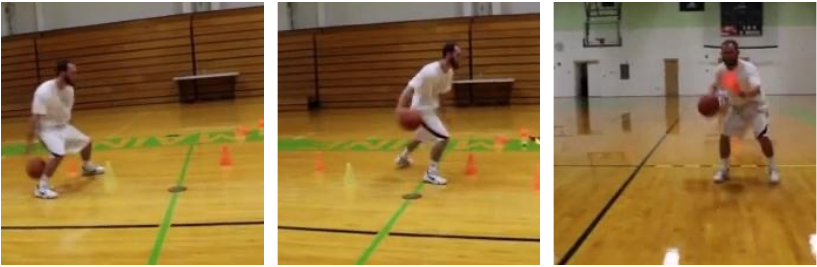}
    \caption{Original video.}
    \label{fig:mask_comparison_original_video}
\end{subfigure}
\begin{subfigure}[t]{.49\linewidth}
    \centering
    \includegraphics[width=0.99\textwidth]{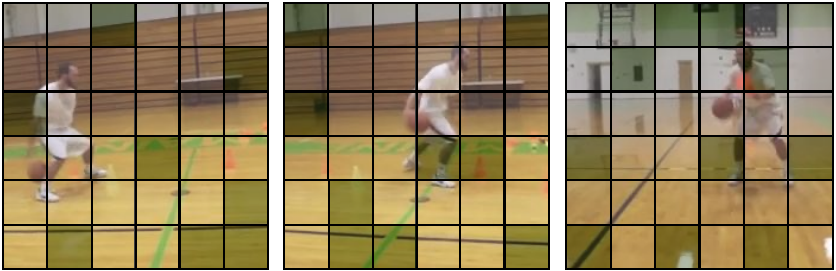}
    \caption{Random mask.}
    \label{fig:mask_comparison_random}
\end{subfigure}
\begin{subfigure}[t]{.49\linewidth}
    \centering
    \includegraphics[width=0.99\textwidth]{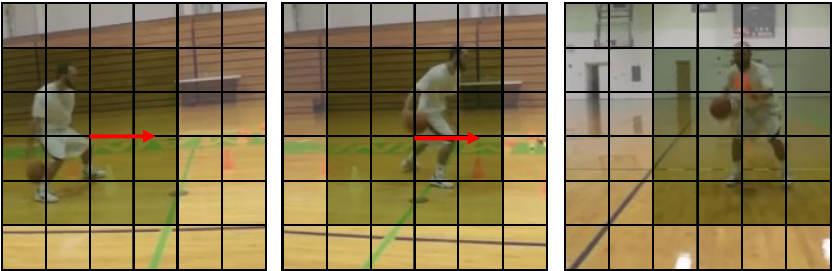}
    \caption{Motion-guided mask.}
    \label{fig:mask_comparison_motion}
\end{subfigure}
\begin{subfigure}[t]{.49\linewidth}
    \centering
    \includegraphics[width=0.99\textwidth]{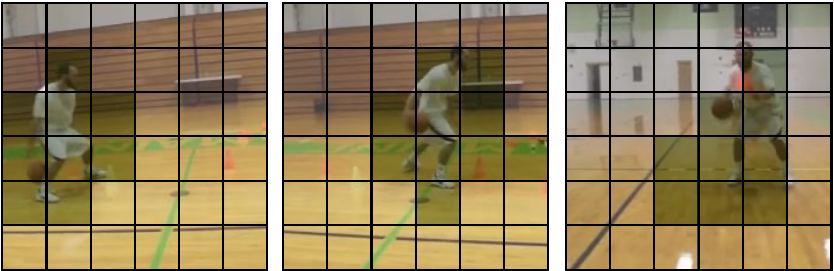}
    \caption{Text-guided mask. ``A man dribbling a basketball''}
    \label{fig:mask_comparison_text}
\end{subfigure}
\vspace{-2mm}
\caption{Illustration of different masking strategies. \ref{fig:mask_comparison_random}: Random masking~\cite{Feichtenhofer2022MaskedAA, Tong2022VideoMAEMA} randomly masks patches independently of their contents. \ref{fig:mask_comparison_motion}: Motion-guided masking~\cite{fan2023motion, huang2023mgmae} tracks the motion of patches over time to mask a moving volume. \ref{fig:mask_comparison_text}: Our proposed text-guided masking masks the top video patch-to-text correspondence.}
\vspace{-2mm}
\end{figure}

The success of masked language modeling~\cite{devlin2018bert, liu2019roberta} has recently inspired the adoption of the masked autoencoder (MAE) for masked image and video modeling. Masking out random image patches~\cite{he2022masked} and reconstructing the missing image patches via an asymmetric encoder-decoder architecture achieves promising results in image recognition. In a similar fashion, works such as VideoMAE~\cite{Tong2022VideoMAEMA} and ST-MAE~\cite{Feichtenhofer2022MaskedAA} achieve promising results in video recognition by extending random masking from 2D image patches to 3D video cubes.

These initial works demonstrate the strong potential of masked visual modeling. Subsequent works have further explored the question of ``where to mask?'' While simple and effective, random masking assumes that input information density is uniformly distributed. Several new masking strategies have been proposed which challenge this assumption and attempt to directly mask visual saliency. For instance, in image domain, SemMAE~\cite{li2022semmae} decomposes foreground into coarse semantic parts and then masks visual patches from each semantic segment according to a defined sampling probability. AutoMAE~\cite{chen2023improving} utilizes adversarial training with bounding boxes to learn an object-centric mask. Both of these works intend to mask the foreground more often than background. These masking algorithms achieve better performance on image recognition benchmarks than the random masking baseline. However, there exists a trade-off point: masking information dense regions too aggressively degrades performance on downstream tasks~\cite{li2022semmae}.

In video domain, MGM~\cite{fan2023motion} and MGMAE~\cite{huang2023mgmae} leverage motion as a video-specific prior for saliency. Specifically, they mask the patches with highest motion over time, where motion is obtained either from motion vectors in the video codec~\cite{fan2023motion} or optical flow~\cite{huang2023mgmae}. Motion-guided masking achieves better performance than random masking in video domain, suggesting that masking saliency is also important for masked video modeling.

These previous works exploit various visual priors with the goal of masking objects and motion. However, their robustness depends on how often input videos match the underlying statistical assumptions. For instance, not all videos have higher foreground motion than background motion. On the other hand, natural language captions are an information dense representation of video that describe both  ``nouns'' (e.g. humans and objects) and ``verbs'' (e.g. actions), without the need to make any prior assumptions. Provided a well-aligned vision-language model, it is possible to directly mask salient regions according to the input text. This makes natural language a promising source of saliency for masked video modeling that has not yet been explored. Thus, this work explores a new direction of improving masked video modeling with natural language, and presents a strong baseline composed a novel masking algorithm and additional loss.


First, we discard visual priors used in previous works and ask whether the content within captions can already capture the most salient regions of video. To that end, we first introduce a novel text-guided masking (\OURMETHOD{}) algorithm, which masks the video regions with highest correspondence to a given caption that is either machine-generated or human-annotated. We present the interesting insight that text-guided masking is competitive with state-of-the-art motion-guided masking, despite using no explicit motion guidance. This confirms the intuition that captions can capture video saliency without prior assumptions.


Second, to further leverage the semantics of natural language for masked video modeling, we introduce a general unified framework for MAE and masked video-text contrastive learning. For any given masking algorithm, introducing an optional contrastive loss aligns the masked encoder representation with the text. To our knowledge, we are the first to unify the generative nature of MAE pretraining and discriminative nature of masked contrastive learning for video and obtain benefits from both pretraining paradigms.

\begin{figure*}[t]
    \centering
    \includegraphics[width=0.98\textwidth]{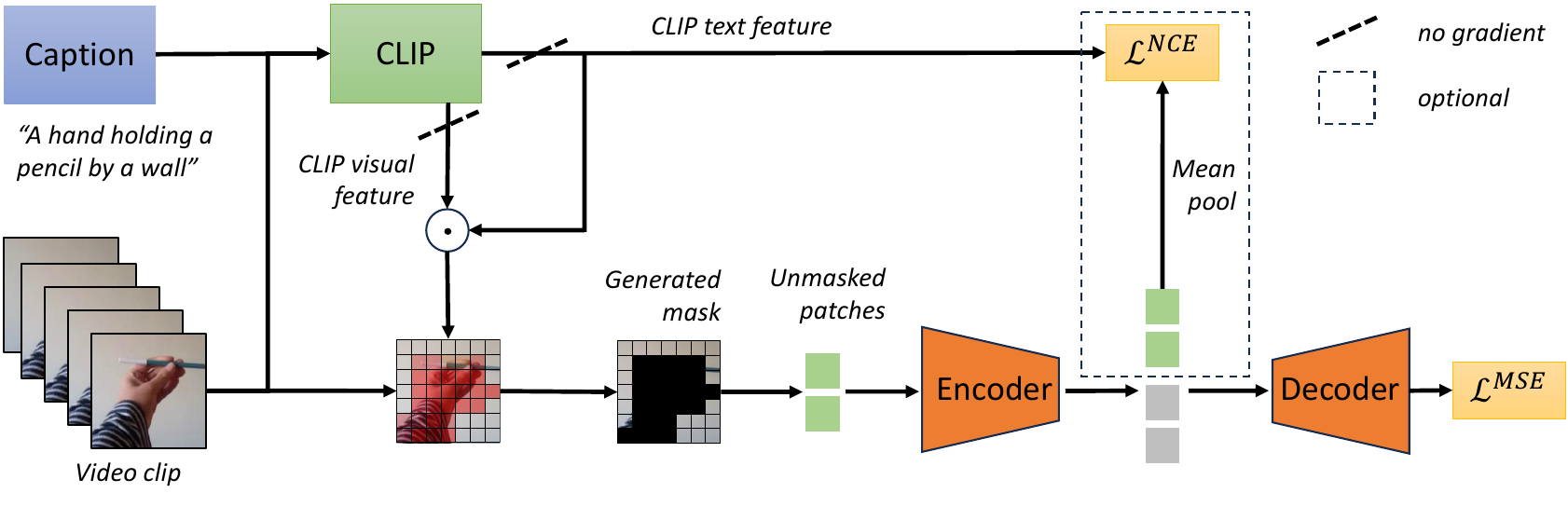}
    \vspace{-1mm}
    \caption{For each video, we generate a caption using an off-shelf image captioning model such as BLIP~\cite{li2022blip}. We then leverage the aligned representation space of CLIP~\cite{radford2021learning} to mask the patches with highest correspondence to the text. The MAE pipeline is identical to VideoMAE~\cite{Tong2022VideoMAEMA}, where the encoder processes the visible patches and the decoder processes the union of encoded visible patches and mask tokens. We additionally introduce an \textit{optional} contrastive loss to align the encoded visible patches with the text. This facilitates semantic-aware reconstruction. BLIP and CLIP receive no gradients.}
    \label{fig:teaser}
    \vspace{-4mm}
\end{figure*}

Training from scratch without any bells and whistles, our text-guided masking outperforms MGM by up to 1.3\% on Kinetics-400 (K400) and 0.5\% on Something-Something V2 (SSv2) in finetuning performance, and by up to 1.7\% in linear evaluation. We show that text-guided masking generalizes to smaller action recognition datasets as well as egocentric action recognition. Lastly, we demonstrate that the synergistic nature of masked contrastive learning also applies to other masking algorithms such as random and motion-guided masking.

In summary, our contributions are:
\begin{enumerate}
    \item Text-guided masking (\OURMETHOD{}) -- a simple yet effective masking algorithm.
    \item Unifying masked video modeling and video-text contrastive learning.
    \item Empirical evidence for the synergistic nature of masked contrastive learning and masked video modeling on five action recognition datasets and one egocentric understanding dataset.
    \item Introducing a new area of research into language-guided masked video modeling and detailed insights that will help inspire future work.
\end{enumerate}

%% file: 2_relatedwork_cameraready.tex
\section{Related Work}


\subsection{Masked Image Modeling}
Early work such as iGPT\cite{chen2020generative} performs masked image modeling at the pixel-level. BEIT~\cite{bao2021beit} elevates the reconstruction target from individual pixels to pretrained dVAE tokens. MAE~\cite{he2022masked} reconstructs normalized pixel patches instead and demonstrates the efficacy of an asymmetric encoder-decoder design with high masking ratio. Recent work addresses the question of whether random masking is ideal. SemMAE~\cite{li2022semmae} uses attention maps to obtain a coarse segmentation and masks patches based on a sampling distribution defined over each semantic class. AutoMAE~\cite{chen2023improving} utilizes adversarial training with bounding boxes to learn an object-centric mask. Both SemMAE and AutoMAE find that masking higher proportion of foreground tokens improves image representations, up until a certain point where masking too aggressively degrades performance. 

In contrast to these works, our work is designed for video and leverages textual information as the primary proxy for video saliency.

\subsection{Masked Video Modeling}
Some works use tokenization-based reconstruction targets. VIMPAC and BEVT use pretrained VQ-VAE~\cite{van2017neural} tokens. However, these works require extra pretraining. MaskFeat instead uses HoG features. VideoMAE~\cite{Tong2022VideoMAEMA} and ST-MAE~\cite{Feichtenhofer2022MaskedAA} directly reconstruct randomly masked 3D video patches, achieving promising results on video benchmarks. Recent work explores whether motion-based priors can lead to improved masking algorithms for video. MGM~\cite{fan2023motion} and MGMAE~\cite{huang2023mgmae} mask the video patches with highest motion, under the assumption that higher motion co-occurs with higher saliency. These motion-guided masks enhance video representations compared to random masking. In contrast to these works, our work leverages textual information as a guide for where to mask, and also introduces a masked video-text contrastive loss that has not been explored yet by video MAE works to the best of our knowledge.

\subsection{Contrastive Visual Pretraining}
MOCO~\cite{he2020momentum, chen2020improved} and SimCLR~\cite{chen2020simple} introduce contrastive learning as an image representation learning paradigm. The encoder is trained by forcing invariance between the encoded representation for two views of the same image, which are typically generated through image augmentations. In video domain, CVRL~\cite{qian2021spatiotemporal} and $\rho$-MoCo~\cite{feichtenhofer2021large} extend contrastive learning to video domain by sampling two subclips from the same video and applying video augmentation. BRAVE~\cite{recasens2021broaden} and LSTCL~\cite{wang2022long} sample overlapping short and long clips to enforce temporal correspondence. These works all use intra-instance positive samples which are generated from the same image/video either through augmentation or re-sampling.

Other works go beyond intra-instance positives to explore inter-instance positive pair sampling. NNCLR~\cite{dwibedi2021little} uses nearest-neighbor images as positive samples to the anchor. Similarly, IIVCL~\cite{fan2023look} uses multiple nearest-neighbor videos as the positive samples to the anchor.

In contrast to these works, we do video-text contrastive learning rather than visual contrastive learning, and we apply masking on top of the video.

\subsection{Vision-Language Pretraining}
CLIP~\cite{radford2021learning} and related works such as ALIGN~\cite{jia2021scaling} popularized image-to-text contrastive learning on hundreds of millions of image-text pairs. FLIP~\cite{li2023scaling} scales CLIP to higher throughput by introducing image masking. A second line of works such as CoCa~\cite{yu2022coca} and Florence~\cite{yuan2021florence} have explored captioning as a vision-language pretraining task. CoCa additionally introduces an optional image-text contrastive loss. Other works apply masked modeling on both image and text. For example, M3AE~\cite{geng2022m3ae} combines image patches and text tokens as the input to a unified masked autoencoder. In contrast to these works, we unify video MAE and masked video-text contrastive learning, do not use captioning as a task, and do not apply masking to text. We also pretrain from scratch on only $\sim$200K videos. We additionally propose our novel text-guided masking strategy.

Another line of work attempts to recreate the success of CLIP for video-text pretraining. CLIP4CLIP~\cite{luo2022clip4clip} takes pretrained CLIP and applies it frame-wise to video and explores different temporal aggregation strategies such as mean pooling and Transformer encoder to achieve a video-level embedding. CLIP4CLIP only explores retrieval tasks. ViCLIP~\cite{wang2023internvid} upgrades the ViT image encoder from CLIP with spatiotemporal attention to make it a video encoder, and trains on a self-curated high-quality dataset with 200M video-text pairs. InternVideo~\cite{wang2022internvideo} alternates between video MAE pretraining and video-text contrastive learning using a different visual backbone for the contrastive learning. In contrast, our approach jointly optimizes the video MAE and video-text contrastive loss with the same visual backbone and the contrastive loss operates on the masked MAE encoder output. We do not use any cross-attention layers nor multimodal fusion. We additionally introduce our novel text-guided masking algorithm.

%% file: 3_method_cameraready.tex
\section{Method}
\subsection{Revisiting Video Masked Autoencoders}
\noindent Given a video $V \in \mathbb{R}^{T \times H \times W \times C }$, where $T,H,W,C$ denote the number of frames, height, width, and RGB-channels, the video is typically first split into cubes of size $t \times h \times w \times C$ and processed with a patch embedding layer $\mathcal{P}$ to obtain a sequence of cube embeddings $V_p$. Typically $t$ = 2 and $h = w = 16$.

\begin{equation}
    V_p = \mathcal{P}(V); V_p \in \mathbb{R}^{\frac{T}{t} \times \frac{H}{h} \times \frac{W}{w} \times D}
    \label{equ:patch_embedding}
\end{equation}

Next, a masking function $\eta$ (e.g. random~\cite{Feichtenhofer2022MaskedAA}, tube~\cite{Tong2022VideoMAEMA}, or motion-guided~\cite{fan2023motion, huang2023mgmae}) generates a binary mask $M$ to select a set of visible patches with mask ratio $\gamma$.

\begin{equation}
\begin{split}
    M &= \eta(V_p,\gamma)\\
    V_{\textrm{p\_visible}} &= V_p \odot (\sim M) \\
    V_{\textrm{p\_masked}} &= V_p \odot M 
\end{split}
\label{equ:masking_tokens}
\end{equation}

The encoder $\phi$ then processes only the visible patches $V_{\text{p\_visible}}$ while the decoder $\xi$ processes the full set of encoded patches and masked tokens $\phi(V_{\text{p\_visible}}) \cup V_{\text{p\_masked}}$ to reconstruct the video. This work uses the same asymmetric encoder-decoder design as~\cite{Tong2022VideoMAEMA, Feichtenhofer2022MaskedAA, he2022masked}.

\begin{equation}
    E=\phi(V_{\text{p\_visible}}), \quad
    V'=\xi(E \cup V_{\text{p\_masked}})
    \label{equ:mae_encoder_decoder}
\end{equation}

Finally, the model is trained with the MSE reconstruction loss $\mathcal{L}_{\textrm{MSE}}$, which is computed between $V$ and $V'$. In this work, we propose a novel mask generator $\eta(\gamma)$ which is guided by text.

\subsection{Caption Generation}
Because Kinetics-400 and Something-Something v2 do not have human annotated captions, we utilize BLIP-2~\cite{li2022blip, li2023blip} offline to generate video-text pairs for pretraining. For each pretraining video, we uniformly sample 3 keyframes and inference a caption per frame, for a total of 3 captions per video. During pretraining, we randomly sample from these 3 captions per video to form the text-video pair. Note that the captioning model is only used offline to obtain captions and does not receive any gradients during training. Note that using off-shelf captioning models for video is convenient but leads to noisier captions than human annotation. Other caption sources are ablated in~\Cref{tab:supp_text_source_results}.

\subsection{Text-Guided Masking}
We leverage the aligned representation space of CLIP~\cite{radford2021learning} to compute a text-guided mask. First, we compute the visual features. For each frame $f_t$, we compute the feature map $V_t \in \mathbb{R}^{\frac{H}{h} \times \frac{W}{w} \times D}$ using ViT-B/32. This is done by patchifying each frame and resizing each patch to full input resolution. We then take the CLS token as the embedding per patch. To get the text embedding $w \in \mathbb{R}^D$, we follow CLIP and take the activation map from the last layer of the transformer at the [\texttt{EOS}] token. We then compute the cosine similarity between $V_t$ and $w$ and take the top $k$ patches by text-video cosine similarity to form the binary mask $\mathcal{M}_{t}$, where $k = \frac{H}{h} \cdot \frac{W}{w} \cdot \gamma$ and $\sum{\mathcal{M}_{t}} / \ell(\mathcal{M}_{t}) = \gamma$ to satisfy the masking ratio $\gamma$. We then apply $\mathcal{M}_{t}$ per frame $f_t$ to obtain the visible $V_{\text{p\_visible}}$ and masked $V_{\text{p\_masked}}$.

\subsection{Video-Text Alignment}
The video-text contrastive loss is a standalone module that can be optionally applied on top of the MAE pipeline for improved performance. The MAE encoder $\phi$ already processes only the visible patches $V_{\text{p\_visible}}$, so no additional computation from the encoder is required for the video-text contrastive loss, as shown in~\Cref{fig:teaser}. Let $i$ index the mini-batch. The global video embedding $v_{i}$ for video $i$ is computed by mean pooling $\phi(V_{\text{p\_visible}})$ across all patches. We then compute $(\mathcal{L}^{\textrm{NCE}}(v_i, t_i, t_{j \neq i}) + \mathcal{L}^{\textrm{NCE}}(t_i, v_i, v_{j \neq i})) / 2$ over the mini-batch of size $N$, where the negative samples are all other text embeddings $t_{j \neq i}$ and video embeddings $v_{j \neq i}$ respectively. Similar to SimCLR~\cite{chen2020simple, chen2020improved}, we use a prediction head with global batch norm. We note that one convenience of masking is that we get large batch size by design, which is beneficial for contrastive learning.

The InfoNCE loss~\cite{oord2018representation} $\mathcal{L}^{\textrm{NCE}}$ maximizes the similarity of a given sample $q$ with its positive key $k^{+}$, while minimizing similarity to negative samples $\mathcal{N}^{-}$:

\begin{equation}
\label{infonce_loss}
\mathcal{L}^{\textrm{NCE}}(q, k^+,\ \mathcal{N}^-) = -\textrm{log} \frac{\textrm{exp}(sim(q, k^+) / \tau)}{\sum\limits_{k \in \{k^+\} \cup \mathcal{N}^-}\textrm{exp}(sim(q, k) / \tau)}
\end{equation}
\noindent where $\tau > 0$ is a temperature hyper-parameter and $sim(\cdot)$ denotes the similarity function --- which in this work is the dot product (cosine) similarity between two $\ell_2$ normalized vectors: $sim(q, k) = q \cdot k = q^{T}k / ( \lVert q \rVert \lVert k \rVert )$. 

\subsection{Text-Guided MAE}
The final loss is either $\mathcal{L}_{\textrm{MSE}}$ in the case of pure MAE, or $\mathcal{L}_{\textrm{MSE}} + \mathcal{L}^{\textrm{NCE}}$ when MAE and video-text contrastive loss are combined. After this self-supervised pretraining, the model is then transferred to downstream tasks such as classification via finetuning and linear probe with cross-entropy loss.

%% file: 4_results_cameraready.tex
\section{Results}
\label{sec:results}

\subsection{Datasets}
\noindent We conduct experiments on six commonly used datasets:

\noindent\textbf{Something-Something V2 (SSv2)}~\cite{goyal2017something} contains 220K videos with 174 action classes. SSv2 is considered a motion heavy dataset, as most of the labels are defined by the motion and directionality of the actual action. \noindent\textbf{Kinetics-400 (K400)}~\cite{kay2017kinetics} is the de-facto standard dataset used to evaluate video recognition. It contains 240K Internet videos with 400 action classes. \noindent\textbf{UCF101}~\cite{soomro2012ucf101} is a dataset containing 13K Internet short videos with 101 action classes. \noindent\textbf{HMDB51}~\cite{kuehne2011hmdb} is a dataset containing 5K short movie clips from 51 action classes. \noindent\textbf{Diving48}~\cite{li2018resound} contains 18K untrimmed video clips from 48 action classes, all of which are types of dives. We report the top-1 accuracy on the evaluation set for all datasets following standard practices~\cite{feichtenhofer2019slowfast}. Only UCF101 and HMDB51 have multiple split versions; we use split 1. \noindent\textbf{Epic-Kitchens 55}~\cite{Damen2018EPICKITCHENS} contains around 30K egocentric first-person video clips from nearly 3K action classes. Egocentric videos feature heavy occlusion, camera motion, and jitter.

\subsection{Implementation Details}
\noindent\textbf{Model Configuration:}
The default backbone is ViT-Base~\cite{dosovitskiy2020image} with global joint space-time attention. For fair comparison, we use the same input patch size of $2 \times 16 \times 16$ for all models following~\cite{Tong2022VideoMAEMA}.

\noindent\textbf{Pre-Processing:} 
We pretrain with clips of 16 frames sampled at a temporal stride of 4 for K400 and stride of 2 for SSv2 respectively following~\cite{Tong2022VideoMAEMA}. We use a fixed spatial resolution of $224 \times 224$ for all experiments. We apply multi-scale-crop and horizontal flip augmentation by default (flip is not applied to SSv2). We follow~\cite{Tong2022VideoMAEMA} to use AdamW~\cite{loshchilov2018decoupled} optimizer with a base learning rate $1.5e-4$, weight decay of $0.05$, $\beta=[0.9, 0.95]$, and cosine learning rate decay. 

\noindent\textbf{Finetuning and Linear Probe:}
We use the same 16-frame clip for finetuning and multi-view evaluation protocol following standard practice~\cite{feichtenhofer2019slowfast}. We use TSN-style sampling~\cite{wang2016temporal, wang2018temporal} on SSv2 dataset with 2 temporal $\times$ 3 spatial views during test-time following~\cite{Tong2022VideoMAEMA} for fair comparison. For Kinetics-400, UCF101, HMDB51, Diving48, and Epic-Kitchens 55, we use 5 temporal $\times$ 3 spatial views during test-time following~\cite{Tong2022VideoMAEMA} for fair comparison. See the supplementary material for  hyperparameter details which are mostly the same as~\cite{Tong2022VideoMAEMA}.

\begin{table*}[!t]
\begin{center}
    \addtolength{\tabcolsep}{0.2em}
    \begin{tabular}{llclcclcc}
     &&& \multicolumn{3}{c}{\textbf{SSv2}} & \multicolumn{3}{c}{\textbf{K400}}   \\ 
    \cmidrule(lr){4-6} \cmidrule(lr){7-9} 
    \textbf{Mask} & \textbf{Backbone} & \textbf{Epochs} & \textbf{Pretrain} & \textbf{FT} & \textbf{LP} & \textbf{Pretrain} & \textbf{FT} & \textbf{LP}  \\
    \midrule
    Tube{$^\mathsection$}~\cite{Tong2022VideoMAEMA} & ViT-B & $200$ & SSv2 & $66.6$ & $25.7$ & K400 & $78.4$ & $38.1$ \\
    MGM{$^\mathsection$}~\cite{fan2023motion} & ViT-B & $200$ & SSv2 & $67.3$ & $33.0$ & K400 & $79.9$ & $32.1$ \\\hline
    \OURMETHOD{} & ViT-B & $200$ & SSv2 & $67.1$ & $26.2$ & K400 & $79.9$ & $33.8$ \\
    \bottomrule
  \end{tabular}
\end{center}
\caption{Comparison between our text-guided masking to (random) tube and motion-guided masking in pure MAE on Something-Something v2 (SSv2) and Kinetics-400 (K400). FT = finetune, LP = linear probe. $\mathsection$ = our reproduction.}
\label{tab:results_mae_only}
\vspace{-5mm}
\end{table*} 

\subsection{Text-Guided Masking}
We first apply our \OURMETHOD{} to pure MAE (no contrastive learning) on SSv2 and K400 in~\Cref{tab:results_mae_only} and compare to random tube masking~\cite{Tong2022VideoMAEMA} and motion-guided masking~\cite{fan2023motion, huang2023mgmae}. We chose MGM~\cite{fan2023motion} over MGMAE~\cite{huang2023mgmae} due to the higher scalability of motion vectors than optical flow.

On SSv2, our \OURMETHOD{} achieves better finetune and linear probe performance than tube masking. On K400, our \OURMETHOD{} achieves better finetune performance than both motion-guided masking and tube masking, and better linear probe performance than motion-guided masking. We do not claim state-of-the-art results, but instead emphasize the surprising and useful insight that text-guided masking is already competitive with other state-of-the-art masking algorithms --- without leveraging any explicit visual cues for saliency such as motion vectors.

\subsection{MAE with Masked Video-Text Contrastive Learning}
\label{sec:mae_and_contrastive}
Next, we introduce the masked video-text contrastive loss as a mask-agnostic module that can be used both by itself, or optionally combined with MAE. We evaluate downstream finetune and linear probe performance for three masking algorithms on SSv2 in~\Cref{tab:results_mae_with_contrastive}. Recall that previous video MAE works do not explore masked video-text contrastive learning, so our results and insights for random tube masking and motion-guided masking are new.

\begin{table*}[!t]
\begin{center}
    \begin{tabular}{lccccc}
         & \multicolumn{2}{c}{\textbf{Components}} & \multicolumn{2}{c}{\textbf{SSv2}} \\
         \cmidrule(lr){2-3} \cmidrule(lr){4-5}
         \textbf{Mask} & \textbf{MAE} & \textbf{V$\rightarrow$T} & \textbf{FT} & \textbf{Linear} \\
         \midrule
         Tube$_{\textrm{p=0.75}}$ & & \checkmark{} & $44.9$ & $12.9$ \\
         Tube$_{\textrm{p=0.75}}$ & \checkmark{} & & $64.9$ & $20.8$ \\
         Tube$_{\textrm{p=0.75}}$ & \checkmark{} & \checkmark{} & $65.5$ & $33.3$ \\
         \hline
         MGM$_{\textrm{p=0.75}}$ & & \checkmark{} & $47.2$ & $6.6$ \\
         MGM$_{\textrm{p=0.75}}$ & \checkmark{} & & $67.3$ & $33.0$ \\
         MGM$_{\textrm{p=0.75}}$ & \checkmark{} & \checkmark{} & $67.0$ & $\mathbf{37.1}$ \\
         \hline
         \OURMETHOD{}$_{\textrm{p=0.6}}$ & \checkmark{} & & $67.1$ & $26.2$ \\
         \OURMETHOD{}$_{\textrm{p=0.6}}$ & \checkmark{} & \checkmark{} & $\mathbf{67.5}$ & $33.4$ \\
         \bottomrule
    \end{tabular}
\end{center}
\caption{Systematic performance breakdown of pure MAE, pure masked video-text contrastive loss, and unified MAE + video-text contrastive loss on Something-Something v2 (SSv2). FT = finetune, LP = linear probe.}
\label{tab:results_mae_with_contrastive}
\vspace{-3mm}
\end{table*}

\noindent \textbf{Comparing within same masking algorithm.}
In this section, we focus on each masking algorithm individually for an apple-to-apple comparison. First, note that pure masked video-text contrastive learning does not achieve competitive results for any masking algorithm. For instance, with random tube masking, pure masked video-text contrastive learning trails pure MAE with random tube masking by over 20\% in finetune and 8\% in linear probe performance. Second, we note that when MAE and contrastive loss are combined, linear probe performance improves notably compared to pure MAE with the same masking algorithm. For example, this boost is 12.5\% for tube masking, 4.1\% for motion-guided masking, and 7.2\% for our \OURMETHOD{}. Finetune performance also improves by 0.6\% for tube masking and by 0.4\% for \OURMETHOD{}. In the case of motion-guided masking, a small drop of 0.3\% in finetune performance is counterbalanced by a 4.1\% improvement in linear probe performance. Overall, the results indicate that the masked video-text contrastive algorithm is synergistic with MAE pretraining, and that this benefit is general across multiple masking algorithms.

\noindent \textbf{Comparing across different masking algorithms.}
With the combined MAE and contrastive loss, we see that \OURMETHOD{} achieves the highest finetune performance (+0.5\% over MGM and +2.0\% over tube masking), while MGM achieves the highest linear probe performance. However, \OURMETHOD{} still achieves a reasonable trade-off which is competitive with both tube and motion-guided masking. Thus, in subsequent experiments, we primarily focus on our \OURMETHOD{} to evaluate its generalizability to other downstream tasks and yield additional insights.

\begin{table}[h]
\begin{center}
    \begin{tabular}{lcccccc|cc|cc}
     && \textbf{UCF} & \textbf{HMDB} & \textbf{Diving48} & \multicolumn{2}{c}{\textbf{UCF}} & \multicolumn{2}{c}{\textbf{HMDB}} & \multicolumn{2}{c}{\textbf{Diving48}}   \\ 
    \cmidrule(lr){3-5} \cmidrule(lr){6-11}
    \textbf{Mask} & \textbf{V $\rightarrow$ T} & \multicolumn{3}{c}{\textbf{Linear Probe}} & \multicolumn{6}{c}{\textbf{R@\{1,5\}}}  \\
    \midrule
    Tube~\cite{Tong2022VideoMAEMA} &            & $70.1$ & $46.1$ & $11.1$ & $89.2$ & $94.5$ & $62.0$ & $78.1$ & $15.0$ & $42.1$ \\
    \hline
    MGM~\cite{fan2023motion}          &            & $70.9$ & $47.0$ & $10.1$ & $85.1$ & $92.2$ & $56.8$ & $73.6$ & $14.9$ & $39.3$ \\
    \hline
    \OURMETHOD{}                         &  & 67.7 & 41.6 & 11.3 & $85.1$ & $91.7$ & $57.8$ & $73.9$ & $13.8$ & $39.9$ \\
    \OURMETHOD{}                & \checkmark & $87.1$ & $64.3$ & $19.9$ & $97.6$ & $99.1$ & $75.7$ & $87.4$ & $18.0$ & $44.8$ \\
    \bottomrule
  \end{tabular}
\end{center}
\vspace{-1mm}
\caption{\OURMETHOD{} generalizes to downstream datasets in linear probe and zero-shot retrieval settings. All results are pretrained for 200 epochs on K400.}
\label{tab:small_dataset_results}
\vspace{-7mm}
\end{table}

\subsection{Transfer Learning}
\subsubsection{Small Action Recognition Datasets.}
We next evaluate \OURMETHOD{} pretrained on K400 when transferred to smaller action recognition datasets: UCF101~\cite{soomro2012ucf101} (13K videos), HDMB51~\cite{kuehne2011hmdb} (5K videos), and Diving48~\cite{li2018resound} (18K videos). These datasets range greatly in content diversity; UCF101 contains Internet videos, HDMB51 contains cinematic clips, and Diving48 contains sports videos. Motivated by the poor linear probe performance of previous MAE works in both image and video domain~\cite{he2022masked, Tong2022VideoMAEMA, fan2023motion}, we focus on evaluation methods that do not finetune the backbone. Specifically, we use both linear evaluation and zero-shot retrieval to see whether \OURMETHOD{} has learned sufficient semantics to generalize even when limited labeled data is available. First, we compare \OURMETHOD{} without contrastive learning to tube masking and motion-guided masking, and see that \OURMETHOD{} is competitive in both the linear probe and zero-shot retrieval settings. When combined with contrastive learning, \OURMETHOD{} achieves a notable performance boost of up to $22.7\%$ in linear probe over \OURMETHOD{} without contrastive learning, as well as up to a $18\%$ boost in recall@1. Thus, we see that the findings from~\Cref{tab:results_mae_only} and~\Cref{tab:results_mae_with_contrastive} still hold true in this small dataset setting.

\subsubsection{Egocentric Action Recognition.}
To further challenge \OURMETHOD{}, we shift to a new task of egocentric action recognition which is uniquely challenging due to featuring first-person perspectives. Egocentric video contains high occlusion, camera motion, and jitter, so it is an interesting setting for evaluating whether our text-guided masking can still be competitive with motion-guided masking --- despite not leveraging any visual cues. In~\Cref{tab:egocentric_results}, we see that without contrastive learning, our \OURMETHOD{} is competitive with both tube and motion-guided masking and only suffers a minor performance drop. With contrastive learning, \OURMETHOD{} improves by 2.5\% in finetune and 5.7\% in linear probe performance, and also outperforms MGM with contrastive learning. This is surprising since MGM explicitly models motion while our \OURMETHOD{} does not leverage explicit visual cues for where to mask. These results indicate that the semantic alignment from video-text contrastive learning is still synergistic with MAE in this high-motion setting.

\begin{table}[h]
\begin{center}
    \begin{tabular}{lccc}
     && \multicolumn{2}{c}{\textbf{Epic-Kitchens}} \\ 
    \cmidrule(lr){3-4}
    \textbf{Mask} & \textbf{V $\rightarrow$ T} & \textbf{FT} & \textbf{Linear}  \\
    \midrule
    Tube~\cite{Tong2022VideoMAEMA} &            & $35.3$ & $16.3$ \\ \hline
    MGM~\cite{fan2023motion}          &            & $35.2$ & $15.3$ \\
    MGM           & \checkmark & $36.9$ & $20.1$  \\ \hline
    \OURMETHOD{}                         &  & $34.7$ & $14.4$ \\
    \OURMETHOD{}                & \checkmark & $\mathbf{37.2}$ & $\mathbf{20.1}$ \\
    \bottomrule
  \end{tabular}
\end{center}
\caption{Egocentric action recognition performance on Epic-Kitchens 55~\cite{Damen2018EPICKITCHENS}. Models are pretrained for 200 epochs on K400.}
\label{tab:egocentric_results}
\vspace{-4mm}
\end{table}

\subsection{Ablations}
\begin{table}[t]
\centering
    \subfloat[\textbf{Mask ratio}]
    {
    \scalebox{1.0}{
        \begin{tabular}{cc}
            \textbf{Ratio} & \textbf{Finetune} \\
            \midrule
             $0.55$ & $67.1$ \\
             $0.60$ & $\mathbf{67.5}$ \\
             $0.75$ & $66.4$ \\
             \bottomrule
        \end{tabular}
        }
        \label{tab:ablation_mask_ratio}
    }
    \subfloat[\textbf{Top vs. bottom-K sampling.}]
    {   
        \scalebox{1.0}{
            \begin{tabular}{lcc}
                \textbf{Masking} & \textbf{Finetune} & \textbf{Linear} \\
                \midrule
                Bottom-K & 67.2 & 33.0  \\
                Tube~\cite{Tong2022VideoMAEMA} & 65.5 & 33.3 \\
                Top-K & $\mathbf{67.5}$ & $\mathbf{33.4}$ \\
                \bottomrule
            \end{tabular}
        }
        \label{tab:ablation_non_topk}
    }
    \subfloat[\textbf{\# of captions per video}.]
    {
        \scalebox{1.0}{
            \begin{tabular}{ccc}
                \textbf{\# Cap.} & \textbf{Finetune} & \textbf{Linear} \\
                \midrule
                1 & 66.5 & 31.3  \\
                3 & $\mathbf{67.5}$ & $\mathbf{33.4}$ \\
                \bottomrule
            \end{tabular}
        }
        \label{tab:ablation_num_captions}
    }
    \label{tab:ablation_all}
    \caption{Ablations with \OURMETHOD{} pretrained for 200 epochs on SSv2 and evaluated on SSv2. \ref{tab:ablation_mask_ratio}: mask ratio, \ref{tab:ablation_non_topk}: masking bottom vs. top patches by textual similarity, \ref{tab:ablation_num_captions}: number of captioned frames.}
    \vspace{-5mm}
\end{table}

\noindent \textbf{Mask Ratio.} \quad
In~\Cref{tab:ablation_mask_ratio}, we ablate the mask ratio used for \OURMETHOD{} with contrastive learning when pretrained on SSv2 for 200 epochs. The optimal masking ratio for our text-guided mask is 0.6, which is significantly lower than other video MAE works. For example, the optimal mask ratio in VideoMAE~\cite{Tong2022VideoMAEMA}, ST-MAE~\cite{Feichtenhofer2022MaskedAA}, and MGMAE~\cite{huang2023mgmae} is 0.9, while the optimal mask ratio in MGM~\cite{fan2023motion} is 0.75. This suggests that \OURMETHOD{} masks more information dense regions.

\noindent \textbf{Top vs. Bottom-K Textual Similarity.} \quad
In~\Cref{tab:ablation_non_topk}, we ablate the choice to mask the top patches by visual-to-text similarity. When we instead mask the bottom patches with lowest visual-to-text similarity, there is a slight drop in performance in both finetune and linear evaluation. However, even with bottom-K sampling, finetune performance is still better than random masking, while linear probe performance is slightly better. We hypothesis that textual guidance is more structured than random masking, but sampling patches with lower CLIP similarity makes the text-to-video alignment noisier.

\noindent \textbf{Number of Captions.} \quad
In~\Cref{tab:ablation_num_captions}, we ablate the number of captions per video. We sample either the center frame or 3 uniformly spaced frames and observe that both finetune and linear probe performance is better with more diverse captions, however the center frame caption already provides strong performance.

\subsection{Visualizations}
We offer visualizations from three perspectives. First, we visualize the masking algorithm and see that the text-guided mask masks the most salient regions of the video matching the natural language description generated by BLIP for the video. Second, we see that the reconstruction quality is decent, meaning the model has learned to solve the reconstruction task. Third, we plot the encoder attention map using the center patch of the center frame as the query. The encoder roughly attends the salient regions of the video. Overall, this suggests that our model has achieved good alignment with the text while solving the MAE reconstruction task. We emphasize that visualizations are not intended to provide a formal explanation for model behavior. Our intention is to provide additional insights into the model to complement our quantitative results.

\begin{figure*}[t]
    \centering
    \includegraphics[width=0.8\textwidth]{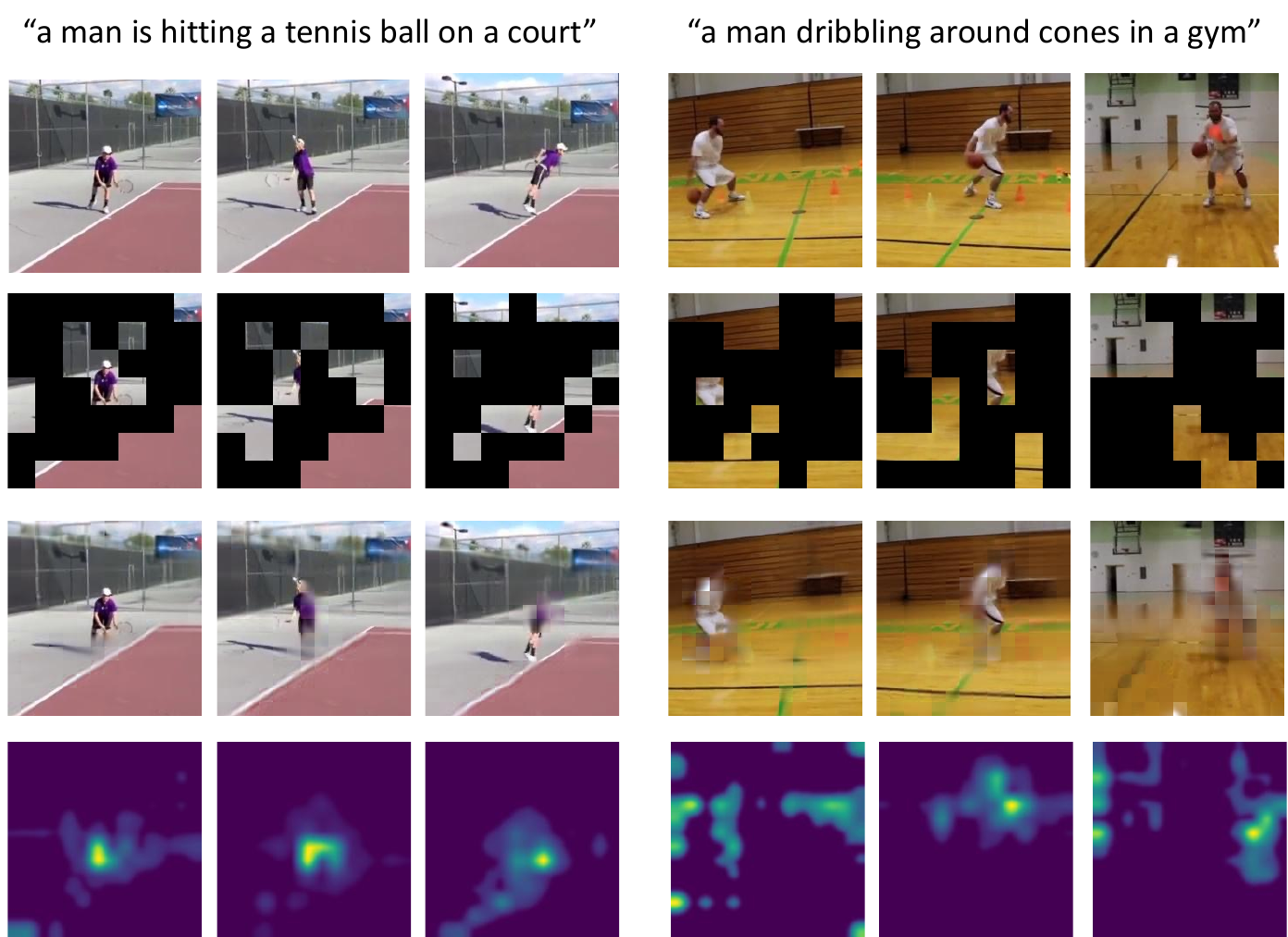}
    \caption{Visualizations from three perspectives: the visualized mask (row 2), reconstructed RGB output (row 3), and encoder attention map (row 4). Our \OURMETHOD{} learns the reconstruction task reasonably well and attends to the salient video regions.}
    \label{fig:visualization}
    \vspace{-3mm}
\end{figure*}

\section{Discussion and Limitations}
\noindent \textbf{Unifying MAE and Contrastive Learning.} \quad
The performance of previous contrastive learning works such as CLIP~\cite{radford2021learning} may make the boost from contrastive learning observed in~\Cref{tab:results_mae_with_contrastive},~\ref{tab:small_dataset_results},~\ref{tab:egocentric_results} seem obvious. We offer two perspectives for why these results are insightful.

First, previous contrastive learning works leverage pretrained weights and/or are trained on hundreds of millions of image/video-text pairs -- which is several orders of magnitude more data than what we use in this paper (roughly 200K video-text pairs). For instance, CLIP~\cite{radford2021learning} and FLIP~\cite{li2023scaling} train on 400 million image-text pairs and ViCLIP~\cite{wang2023internvid} is trained on 200 million video-text pairs. We showed in~\Cref{tab:results_mae_with_contrastive} that on Kinetics-400 and Something-Something, pure video-text contrastive learning achieves much lower performance than pure MAE. The benefits of video-text contrastive learning when training from scratch on our scale of data are only realized when combined with MAE.

Second, careful design is required to unify the generative nature of MAE and discriminative nature of contrastive learning. Previous work even suggests that MAE and video-text contrastive learning are antagonistic, which conflicts with our findings. For instance, FLIP~\cite{li2023scaling} reports degraded finetuning performance when combining image MAE with masked contrastive image-text learning. It is not obvious that MAE and video-text contrastive learning are synergistic for video. Our empirical results contribute the useful insight that this is not only the case, but also that this benefit can be realized across multiple masking algorithms -- even with noisy machine-generated captions on ``regular'' sized datasets.

\begin{wrapfigure}{r}{0.45\textwidth}
    \vspace{-10mm}
    \begin{centering}
        \includegraphics[width=0.45\textwidth]{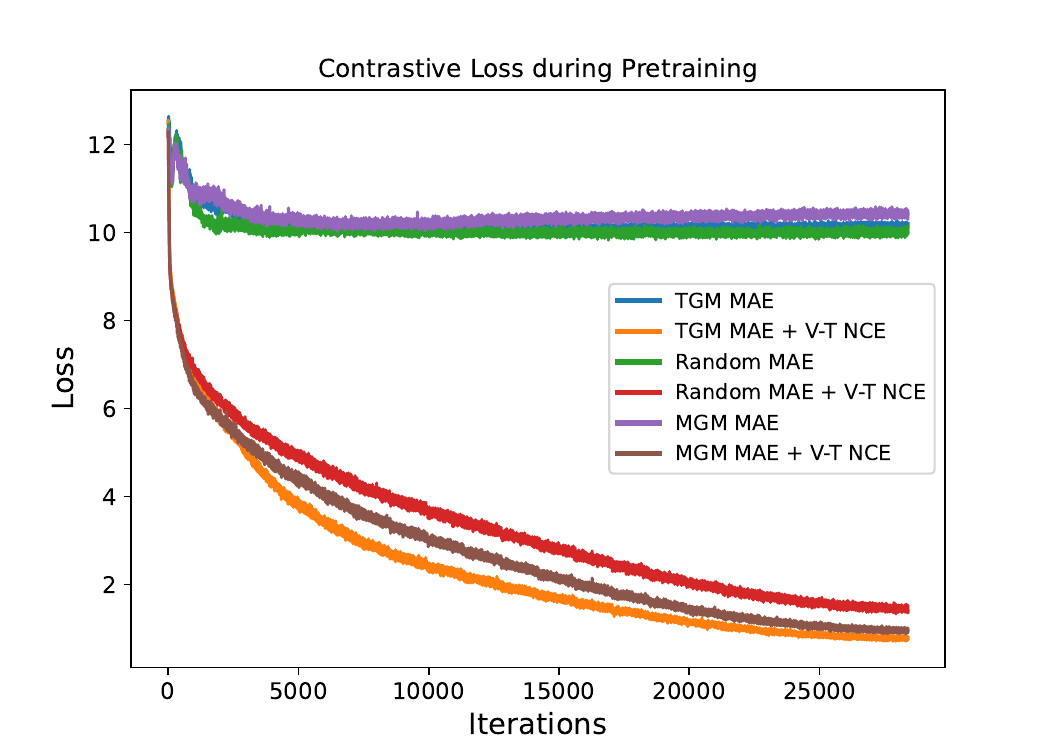}
    \end{centering}
    \vspace{-6mm}
    \caption{Contrastive loss for each mask alg. with and without optimization.}
    \label{fig:contrastive_loss}
    \vspace{-8mm}
\end{wrapfigure}

We leave better theoretical understanding of why these paradigms are complementary to future work. To provide an additional perspective, we plot the contrastive loss with and without optimization of the contrastive loss for all three mask algorithms in~\Cref{fig:contrastive_loss}. We observe that even for pure MAE, the contrastive loss naturally decreases. This suggests that the MAE encoder already learns semantics that somewhat align with text, even when there is no textual supervision. This further motivates the benefit of unifying MAE and contrastive learning to facilitate semantic-aware reconstruction.

\noindent \textbf{Choice of Captions.} \quad
Our work utilizes frame-wise captions from BLIP-2~\cite{li2022blip, li2023blip} which is an off-shelf image captioning model. An interesting question is whether our performance is dependent on this particular captioning model, and whether higher model capacity helps. In~\Cref{tab:supp_text_source_results}, we first test an oracle captioner that directly outputs the action label. We then use GPT3.5~\cite{brown2020language} which is a large language model with several orders of magnitude more parameters than BLIP. GPT3.5 is a pure language model so it requires a textual prompt. Previous works~\cite{pratt2023does, menon2023visual} demonstrate that GPT-3 is capable of generating detailed descriptions about specific object and action categories with only textual inputs. Following CuPL~\cite{pratt2023does}, we utilize three prompt templates listed in~\Cref{tab:supp_text_prompts} to generate ``vision-free'' captions per action class. We then utilize these vision-free captions for pretraining to assess the effect of large-language model generated captions in our combined framework.

We see that linear probe performance for both the oracle and GPT3.5 is much higher than BLIP, while finetune performance drops. We posit that this is primarily because the oracle and GPT3.5 are vision-free, so the generated caption is not guaranteed to capture the granular visual details of any video. However, the textual caption generated by GPT3.5 may abstractly match the action class better since it is asked to describe each action class in general, and GPT3.5 has several orders of magnitude more parameters and training data. In summary, more model capacity and external domain knowledge does not necessarily translate to better performance in our framework, however vision-free captions are still surprisingly useful. Combining BLIP and GPT3.5 captions degrades performance probably due to the conflicting nature of vision-free captions.

\begin{table}[t]
\centering
    \subfloat[\textbf{}]
    {
        \scalebox{1.0}{
            \begin{tabular}{l}
                \textbf{GPT Prompts} \\
                \midrule
                "Describe the action \{\}." \\
                "What does a person \{\} look like?" \\
                "What does the act of \{\} look like?" \\
                \bottomrule
            \end{tabular}
        }
        \label{tab:supp_text_prompts}
    }\hfill
    \subfloat[\textbf{}]
    {
        \scalebox{1.0}{
            \begin{tabular}{lcc}
                \textbf{Text Source} & \textbf{FT} & \textbf{Linear} \\
                \midrule
                Oracle & 64.1 & 51.5 \\
                GPT3.5~\cite{brown2020language} & 66.2 & \textbf{54.0} \\ 
                BLIP~\cite{li2023scaling} & \textbf{67.5} & 33.4 \\
                BLIP + GPT3.5 & 65.8 & 51.4 \\
                \bottomrule
            \end{tabular}
        }
        \label{tab:supp_text_source_results}
    }
    \label{tab:supp_text_source_ablation}
    \vspace{-1mm}
    \caption{\ref{tab:supp_text_prompts}: Prompts provided to GPT3.5 to generate text-based captions per action. Inspired by CuPL~\cite{pratt2023does}. \ref{tab:supp_text_source_results}: Effect of different text sources when pretraining on SSv2 for 200 epochs and evaluated on SSv2.}
    \vspace{-5mm}
\end{table}

\noindent \textbf{Computational Efficiency.} \quad
Although mask generation requires CLIP inference, the wall clock time for training is not significantly higher. This is because the high mask ratio and asymmetric encoder-decoder design of the MAE pipeline contribute to dataloading itself being the primary bottleneck~\cite{Feichtenhofer2022MaskedAA}. BLIP and CLIP both are frozen.

\noindent \textbf{Limitations.} \quad
One limitation is the reliance upon captions. However, the captions used in our work are far from perfect. For instance, BLIP is an image captioning model and frame-wise captioning may not capture temporal details of video. Despite this limitation, we showed the efficacy of these noisy captions within our framework. Video captioning is a difficult problem and we expect that as state-of-the-art improves, our performance would also improve. Another limitation is that strong vision-text pretraining is needed to leverage captions for mask generation. However, the growing availability of CLIP-like models makes research into how to leverage these resources all the more timely, especially for designing real-world systems. We have proposed both a new direction of research and a simple yet effective baseline that offers room for further research.

%% file: 5_conclusion_cameraready.tex
\section{Conclusion}
We motivated a new research direction into leveraging natural language to improve masked video representation learning. We first presented \OURMETHOD{}, a novel masking algorithm that masks the regions of video with highest alignment to text captions. Next, we introduced masked video-to-text contrastive learning as an optional module that can be combined with video MAE to enrich semantic learning across a host of masking algorithms. When combined with specifically our \OURMETHOD{}, we observe improved performance in finetuning, linear probe, and even zero-shot retrieval across six different downstream datasets. Our approach is simple yet effective and sets a baseline for future research in this new direction of language-guided MAE.

\noindent \textbf{Acknowledgements.} \quad We thank Linda Liu and Pichao Wang for their valuable feedback on this work.

%% file: supplementary.tex
\section{Additional Results}
\subsection{More Epochs}
In~\Cref{tab:supp_more_epochs} we display the results with more epochs of pretraining on unlabeled Kinetics-400. We see that both finetuning and linear performance both continue to improve with more epochs of pretraining, with linear evaluation achieving higher delta of improvement with longer pretraining. Performance does not saturate with longer pretraining.

\begin{table}[t]
    \centering
    \begin{tabular}{lccc}
             \toprule
             & & \multicolumn{2}{c}{\textbf{Top-1 Acc}} \\
             \cmidrule(lr){2-3}
             Mask & Epochs & FT & Linear \\
             \hline
             \OURMETHOD{}$_{\textrm{p=0.6}}$ & 200 & 79.6 & 59.4 \\
              & 400 & 80.2 & 62.4 \\
              & 800 & 80.3 & 63.4 \\
             \bottomrule
        \end{tabular}
        \caption{Results with more epochs of pretraining on unlabeled K400. Results are finetuning (FT) and linear evaluation on K400.}
    \label{tab:supp_more_epochs}
\end{table}

\section{Additional Visualizations}
\begin{figure*}[t]
    \centering
    \includegraphics[width=0.98\textwidth]{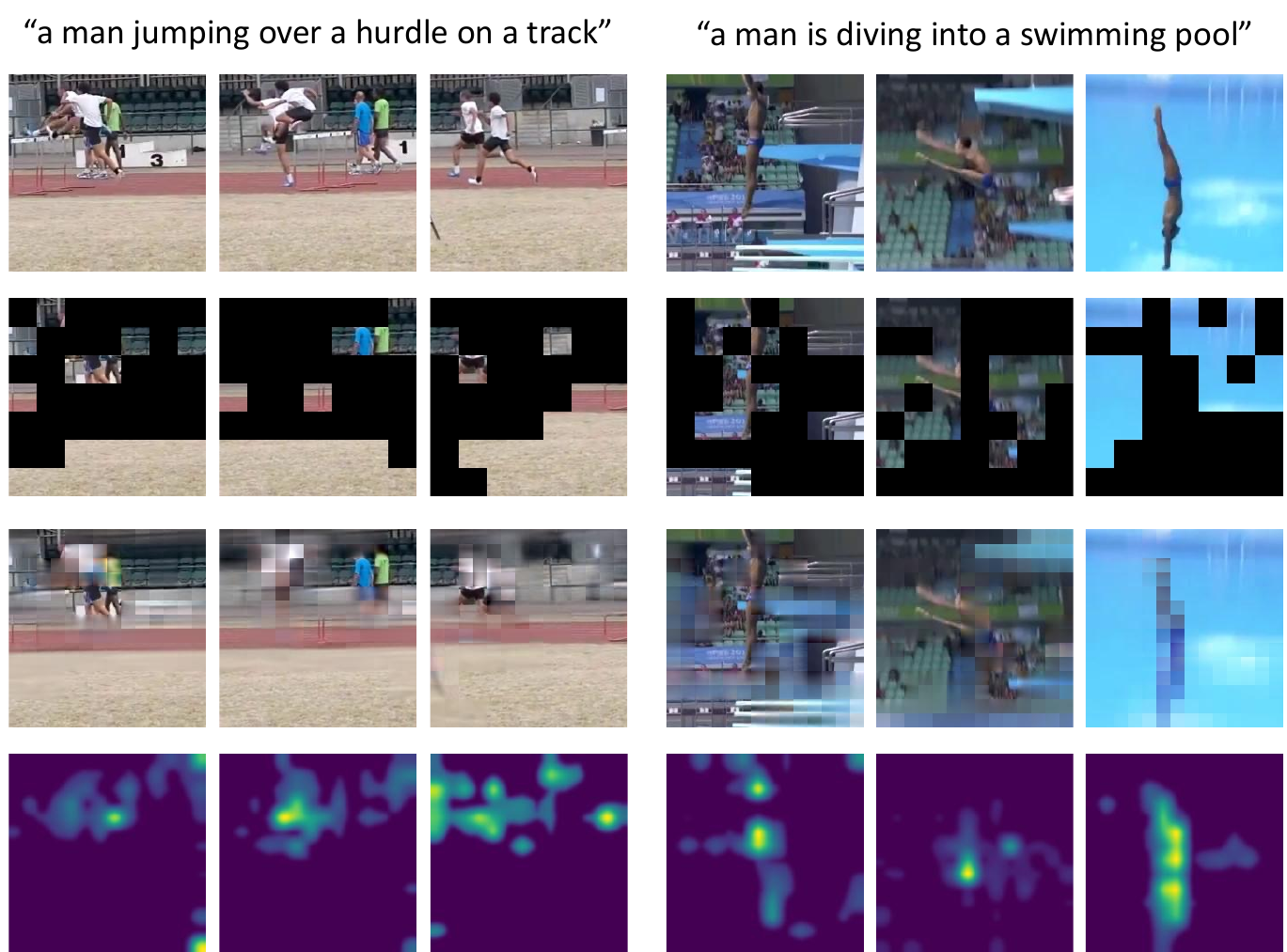}
    \caption{Additional visualizations from three perspectives: the visualized mask, reconstructed RGB output, and encoder attention map. We see that our \OURMETHOD{} solves the reconstruction task reasonably well and learns to attend to the salient regions of the video.}
    \label{fig:supp_additionalvis}
\end{figure*}

In~\Cref{fig:supp_additionalvis} we provide additional visualizations. The model both masks and attends to the most salient regions of video that correspond to the provided natural language description. Again, we emphasize that visualizations are not intended to provide a formal explanation for model behavior. Our intention is to provide additional insights into the model to complement our quantitative results.

\section{Additional Hyperparameters}
We mostly follow the same hyperparameters as~\cite{Tong2022VideoMAEMA}. ~\Cref{table:hyperparameters_pretraining} and \Cref{table:hyperparameters_finetuning} show the configurations for pretraining and finetuning.

\begin{table}[!th]
\small
\begin{center}
    \begin{tabular}{l|cc}
    config & SSv2 & K400 \\
    \hline
    optimizer & \multicolumn{2}{c}{AdamW} \\
    base learning rate$^{\dagger}$ & \multicolumn{2}{c}{1.5e-4} \\
    weight decay & \multicolumn{2}{c}{0.05} \\
    optimizer momentum & \multicolumn{2}{c}{$\beta_1, \beta_2=0.9, 0.95$} \\
    batch size & \multicolumn{2}{c}{512} \\
    learning rate schedule & \multicolumn{2}{c}{cosine decay~\cite{loshchilov2016sgdr}} \\
    warmup epochs & \multicolumn{2}{c}{40} \\
    flip augmentation & no & yes \\
    augmentation & \multicolumn{2}{c}{MultiScaleCrop}
  \end{tabular}
\end{center}
\vspace{-2mm}
\caption{Pretraining hyperparameters. $^{\dagger}:$ we follow the linear LR scaling rule. $lr=base\_lr \times batch\_size / 256$.}
\label{table:hyperparameters_pretraining}
\end{table}

\begin{table}[!th]
\small
\begin{center}
    \begin{tabular}{l|cccccc}
    config & SSv2 & K400 \\
    \hline
    optimizer & \multicolumn{2}{c}{AdamW} \\
    base learning rate & 5e-4 & 1e-3 \\
    weight decay & \multicolumn{2}{c}{0.05} \\
    optimizer momentum & \multicolumn{2}{c}{$\beta_1, \beta_2=0.9, 0.999$} \\
    layer-wise lr decay & 0.75~\cite{bao2021beit} & 0.75 \\
    batch size & \multicolumn{2}{c}{384} \\
    learning rate schedule & \multicolumn{2}{c}{cosine decay} \\
    repeated augmentation & 2~\cite{hoffer2020augment} & 2 \\
    warmup epochs & 5 & 5 \\
    total epochs & 30 & 75 \\
    flip augmentation & no & yes \\
    drop path & 0.1 & 0.1
  \end{tabular}
\end{center}
\caption{Finetuning hyperparameters.}
\label{table:hyperparameters_finetuning}
\end{table}



%% file: main.bbl
\begin{thebibliography}{10}
\providecommand{\url}[1]{\texttt{#1}}
\providecommand{\urlprefix}{URL }
\providecommand{\doi}[1]{https://doi.org/#1}

\bibitem{bao2021beit}
Bao, H., Dong, L., Piao, S., Wei, F.: Beit: Bert pre-training of image transformers. In: International Conference on Learning Representations (2021)

\bibitem{brown2020language}
Brown, T., Mann, B., Ryder, N., Subbiah, M., Kaplan, J.D., Dhariwal, P., Neelakantan, A., Shyam, P., Sastry, G., Askell, A., et~al.: Language models are few-shot learners. Advances in neural information processing systems  \textbf{33},  1877--1901 (2020)

\bibitem{chen2023improving}
Chen, H., Zhang, W., Wang, Y., Yang, X.: Improving masked autoencoders by learning where to mask. arXiv preprint arXiv:2303.06583  (2023)

\bibitem{chen2020generative}
Chen, M., Radford, A., Child, R., Wu, J., Jun, H., Luan, D., Sutskever, I.: Generative pretraining from pixels. In: International conference on machine learning. pp. 1691--1703. PMLR (2020)

\bibitem{chen2020simple}
Chen, T., Kornblith, S., Norouzi, M., Hinton, G.: A simple framework for contrastive learning of visual representations. In: International conference on machine learning. pp. 1597--1607. PMLR (2020)

\bibitem{chen2020improved}
Chen, X., Fan, H., Girshick, R., He, K.: Improved baselines with momentum contrastive learning. arXiv preprint arXiv:2003.04297  (2020)

\bibitem{Damen2018EPICKITCHENS}
Damen, D., Doughty, H., Farinella, G.M., Fidler, S., Furnari, A., Kazakos, E., Moltisanti, D., Munro, J., Perrett, T., Price, W., Wray, M.: Scaling egocentric vision: The epic-kitchens dataset. In: European Conference on Computer Vision (ECCV) (2018)

\bibitem{devlin2018bert}
Devlin, J., Chang, M.W., Lee, K., Toutanova, K.: Bert: Pre-training of deep bidirectional transformers for language understanding. arXiv preprint arXiv:1810.04805  (2018)

\bibitem{dosovitskiy2020image}
Dosovitskiy, A., Beyer, L., Kolesnikov, A., Weissenborn, D., Zhai, X., Unterthiner, T., Dehghani, M., Minderer, M., Heigold, G., Gelly, S., et~al.: An image is worth 16x16 words: Transformers for image recognition at scale. In: International Conference on Learning Representations (2020)

\bibitem{dwibedi2021little}
Dwibedi, D., Aytar, Y., Tompson, J., Sermanet, P., Zisserman, A.: With a little help from my friends: Nearest-neighbor contrastive learning of visual representations. In: Proceedings of the IEEE/CVF International Conference on Computer Vision. pp. 9588--9597 (2021)

\bibitem{fan2023motion}
Fan, D., Wang, J., Liao, S., Zhu, Y., Bhat, V., Santos-Villalobos, H., MV, R., Li, X.: Motion-guided masking for spatiotemporal representation learning. In: Proceedings of the IEEE/CVF International Conference on Computer Vision. pp. 5619--5629 (2023)

\bibitem{fan2023look}
Fan, D., Yang, D., Li, X., Bhat, V., Rohith, M.: Look globally and locally: Inter-intra contrastive learning from unlabeled videos. In: ICLR 2023 Workshop on Mathematical and Empirical Understanding of Foundation Models (2023)

\bibitem{feichtenhofer2019slowfast}
Feichtenhofer, C., Fan, H., Malik, J., He, K.: Slowfast networks for video recognition. In: Proceedings of the IEEE/CVF international conference on computer vision. pp. 6202--6211 (2019)

\bibitem{feichtenhofer2021large}
Feichtenhofer, C., Fan, H., Xiong, B., Girshick, R., He, K.: A large-scale study on unsupervised spatiotemporal representation learning. In: Proceedings of the IEEE/CVF Conference on Computer Vision and Pattern Recognition. pp. 3299--3309 (2021)

\bibitem{Feichtenhofer2022MaskedAA}
Feichtenhofer, C., Li, Y., He, K., et~al.: Masked autoencoders as spatiotemporal learners. Advances in neural information processing systems  \textbf{35},  35946--35958 (2022)

\bibitem{geng2022m3ae}
Geng, X., Liu, H., Lee, L., Schuurmans, D., Levine, S., Abbeel, P.: M3ae: Multimodal masked autoencoders learn transferable representations. Tech. rep., Technical Report (2022)

\bibitem{goyal2017something}
Goyal, R., Ebrahimi~Kahou, S., Michalski, V., Materzynska, J., Westphal, S., Kim, H., Haenel, V., Fruend, I., Yianilos, P., Mueller-Freitag, M., et~al.: The" something something" video database for learning and evaluating visual common sense. In: Proceedings of the IEEE international conference on computer vision. pp. 5842--5850 (2017)

\bibitem{he2022masked}
He, K., Chen, X., Xie, S., Li, Y., Doll{\'a}r, P., Girshick, R.: Masked autoencoders are scalable vision learners. In: Proceedings of the IEEE/CVF conference on computer vision and pattern recognition. pp. 16000--16009 (2022)

\bibitem{he2020momentum}
He, K., Fan, H., Wu, Y., Xie, S., Girshick, R.: Momentum contrast for unsupervised visual representation learning. In: Proceedings of the IEEE/CVF conference on computer vision and pattern recognition. pp. 9729--9738 (2020)

\bibitem{hoffer2020augment}
Hoffer, E., Ben-Nun, T., Hubara, I., Giladi, N., Hoefler, T., Soudry, D.: Augment your batch: Improving generalization through instance repetition. In: Proceedings of the IEEE/CVF Conference on Computer Vision and Pattern Recognition. pp. 8129--8138 (2020)

\bibitem{huang2023mgmae}
Huang, B., Zhao, Z., Zhang, G., Qiao, Y., Wang, L.: Mgmae: Motion guided masking for video masked autoencoding. In: Proceedings of the IEEE/CVF International Conference on Computer Vision. pp. 13493--13504 (2023)

\bibitem{jia2021scaling}
Jia, C., Yang, Y., Xia, Y., Chen, Y.T., Parekh, Z., Pham, H., Le, Q., Sung, Y.H., Li, Z., Duerig, T.: Scaling up visual and vision-language representation learning with noisy text supervision. In: International conference on machine learning. pp. 4904--4916. PMLR (2021)

\bibitem{kay2017kinetics}
Kay, W., Carreira, J., Simonyan, K., Zhang, B., Hillier, C., Vijayanarasimhan, S., Viola, F., Green, T., Back, T., Natsev, P., et~al.: The kinetics human action video dataset. arXiv preprint arXiv:1705.06950  (2017)

\bibitem{kuehne2011hmdb}
Kuehne, H., Jhuang, H., Garrote, E., Poggio, T., Serre, T.: Hmdb: a large video database for human motion recognition. In: 2011 International conference on computer vision. pp. 2556--2563. IEEE (2011)

\bibitem{li2022semmae}
Li, G., Zheng, H., Liu, D., Wang, C., Su, B., Zheng, C.: Semmae: Semantic-guided masking for learning masked autoencoders. Advances in Neural Information Processing Systems  \textbf{35},  14290--14302 (2022)

\bibitem{li2023blip}
Li, J., Li, D., Savarese, S., Hoi, S.: Blip-2: Bootstrapping language-image pre-training with frozen image encoders and large language models. arXiv preprint arXiv:2301.12597  (2023)

\bibitem{li2022blip}
Li, J., Li, D., Xiong, C., Hoi, S.: Blip: Bootstrapping language-image pre-training for unified vision-language understanding and generation. In: International Conference on Machine Learning. pp. 12888--12900. PMLR (2022)

\bibitem{li2023scaling}
Li, Y., Fan, H., Hu, R., Feichtenhofer, C., He, K.: Scaling language-image pre-training via masking. In: Proceedings of the IEEE/CVF Conference on Computer Vision and Pattern Recognition. pp. 23390--23400 (2023)

\bibitem{li2018resound}
Li, Y., Li, Y., Vasconcelos, N.: Resound: Towards action recognition without representation bias. In: Proceedings of the European Conference on Computer Vision (ECCV). pp. 513--528 (2018)

\bibitem{liu2019roberta}
Liu, Y., Ott, M., Goyal, N., Du, J., Joshi, M., Chen, D., Levy, O., Lewis, M., Zettlemoyer, L., Stoyanov, V.: Roberta: A robustly optimized bert pretraining approach. arXiv preprint arXiv:1907.11692  (2019)

\bibitem{loshchilov2016sgdr}
Loshchilov, I., Hutter, F.: Sgdr: Stochastic gradient descent with warm restarts. arXiv preprint arXiv:1608.03983  (2016)

\bibitem{loshchilov2018decoupled}
Loshchilov, I., Hutter, F.: Decoupled weight decay regularization. In: International Conference on Learning Representations (2018)

\bibitem{luo2022clip4clip}
Luo, H., Ji, L., Zhong, M., Chen, Y., Lei, W., Duan, N., Li, T.: Clip4clip: An empirical study of clip for end to end video clip retrieval and captioning. Neurocomputing  \textbf{508},  293--304 (2022)

\bibitem{menon2023visual}
Menon, S., Vondrick, C.: Visual classification via description from large language models. In: The Eleventh International Conference on Learning Representations (2023), \url{https://openreview.net/forum?id=jlAjNL8z5cs}

\bibitem{oord2018representation}
Oord, A.v.d., Li, Y., Vinyals, O.: Representation learning with contrastive predictive coding. arXiv preprint arXiv:1807.03748  (2018)

\bibitem{pratt2023does}
Pratt, S., Covert, I., Liu, R., Farhadi, A.: What does a platypus look like? generating customized prompts for zero-shot image classification. In: Proceedings of the IEEE/CVF International Conference on Computer Vision. pp. 15691--15701 (2023)

\bibitem{qian2021spatiotemporal}
Qian, R., Meng, T., Gong, B., Yang, M.H., Wang, H., Belongie, S., Cui, Y.: Spatiotemporal contrastive video representation learning. In: Proceedings of the IEEE/CVF Conference on Computer Vision and Pattern Recognition. pp. 6964--6974 (2021)

\bibitem{radford2021learning}
Radford, A., Kim, J.W., Hallacy, C., Ramesh, A., Goh, G., Agarwal, S., Sastry, G., Askell, A., Mishkin, P., Clark, J., et~al.: Learning transferable visual models from natural language supervision. In: International conference on machine learning. pp. 8748--8763. PMLR (2021)

\bibitem{recasens2021broaden}
Recasens, A., Luc, P., Alayrac, J.B., Wang, L., Strub, F., Tallec, C., Malinowski, M., P{\u{a}}tr{\u{a}}ucean, V., Altch{\'e}, F., Valko, M., et~al.: Broaden your views for self-supervised video learning. In: Proceedings of the IEEE/CVF International Conference on Computer Vision. pp. 1255--1265 (2021)

\bibitem{soomro2012ucf101}
Soomro, K., Zamir, A.R., Shah, M.: Ucf101: A dataset of 101 human actions classes from videos in the wild. arXiv preprint arXiv:1212.0402  (2012)

\bibitem{Tong2022VideoMAEMA}
Tong, Z., Song, Y., Wang, J., Wang, L.: Videomae: Masked autoencoders are data-efficient learners for self-supervised video pre-training. ArXiv  \textbf{abs/2203.12602} (2022)

\bibitem{van2017neural}
Van Den~Oord, A., Vinyals, O., et~al.: Neural discrete representation learning. Advances in neural information processing systems  \textbf{30} (2017)

\bibitem{wang2022long}
Wang, J., Bertasius, G., Tran, D., Torresani, L.: Long-short temporal contrastive learning of video transformers. In: Proceedings of the IEEE/CVF Conference on Computer Vision and Pattern Recognition. pp. 14010--14020 (2022)

\bibitem{wang2016temporal}
Wang, L., Xiong, Y., Wang, Z., Qiao, Y., Lin, D., Tang, X., Van~Gool, L.: Temporal segment networks: Towards good practices for deep action recognition. In: European conference on computer vision. pp. 20--36. Springer (2016)

\bibitem{wang2018temporal}
Wang, L., Xiong, Y., Wang, Z., Qiao, Y., Lin, D., Tang, X., Van~Gool, L.: Temporal segment networks for action recognition in videos. IEEE transactions on pattern analysis and machine intelligence  \textbf{41}(11),  2740--2755 (2018)

\bibitem{wang2023internvid}
Wang, Y., He, Y., Li, Y., Li, K., Yu, J., Ma, X., Li, X., Chen, G., Chen, X., Wang, Y., et~al.: Internvid: A large-scale video-text dataset for multimodal understanding and generation. In: The Twelfth International Conference on Learning Representations (2023)

\bibitem{wang2022internvideo}
Wang, Y., Li, K., Li, Y., He, Y., Huang, B., Zhao, Z., Zhang, H., Xu, J., Liu, Y., Wang, Z., et~al.: Internvideo: General video foundation models via generative and discriminative learning. arXiv preprint arXiv:2212.03191  (2022)

\bibitem{yu2022coca}
Yu, J., Wang, Z., Vasudevan, V., Yeung, L., Seyedhosseini, M., Wu, Y.: Coca: Contrastive captioners are image-text foundation models. arXiv preprint arXiv:2205.01917  (2022)

\bibitem{yuan2021florence}
Yuan, L., Chen, D., Chen, Y.L., Codella, N., Dai, X., Gao, J., Hu, H., Huang, X., Li, B., Li, C., et~al.: Florence: A new foundation model for computer vision. arXiv preprint arXiv:2111.11432  (2021)

\end{thebibliography}
